\documentclass[10pt,twocolumn,letterpaper]{article}

\usepackage[pagenumbers]{cvpr} % To force page numbers, e.g. for an arXiv version

\definecolor{cvprblue}{rgb}{0.21,0.49,0.74}
\usepackage[pagebackref,breaklinks,colorlinks,allcolors=cvprblue]{hyperref}
\usepackage{wrapfig}
\usepackage{multirow}
\usepackage[export]{adjustbox}
\DeclareMathOperator*{\argmax}{arg\,max}

\newcommand{\vae}{CAT\xspace}

\newcommand\freefootnote[1]{%
  \let\thefootnote\relax%
  \footnotetext{#1}%
  \let\thefootnote\svthefootnote%
}
\usepackage[symbol]{footmisc}

\def\paperID{*} % *** Enter the Paper ID here
\def\confName{*}
\def\confYear{*}

\title{CAT: Content-Adaptive Image Tokenization}

\author{
\begin{tabular}{c}
\begin{tabular}{c@{\hskip 0.85cm}c@{\hskip 0.85cm}c}
Junhong Shen$^1$\thanks{Work done during an internship at Meta.} & Kushal Tirumala$^2$ & Michihiro Yasunaga$^2$\\
\end{tabular}\\
\begin{tabular}{c@{\hskip 0.85cm}c@{\hskip 0.85cm}c@{\hskip 0.85cm}c}
Ishan Misra$^2$ & Luke Zettlemoyer$^{2}$ & Lili Yu$^2$\thanks{Joint senior author.}  & Chunting Zhou$^{\dag}$\thanks{Work done while at Meta.}\\
\end{tabular}
\\\\
\normalsize
\begin{tabular}{c}
$^1$ Carnegie Mellon University \\
$^2$ Meta 
\end{tabular}
\\\\
\begin{tabular}{c}
$^1$ \normalsize\texttt{junhongs@andrew.cmu.edu} \\
$^2$ \normalsize\texttt{\{ktirumala, myasu, imisra, lsz, liliyu\}@meta.com}\\
$^\ddagger$ \normalsize\texttt{chunting.violet.zhou@gmail.com}
\end{tabular}
\end{tabular}
}

\begin{document}
\maketitle

\begin{abstract}
Most existing image tokenizers encode images into a fixed number of tokens or patches, overlooking the inherent variability in image complexity. To address this, we introduce \textbf{C}ontent-\textbf{A}daptive \textbf{T}okenizer (\vae), which dynamically adjusts representation capacity based on the image content and encodes simpler images into fewer tokens. 
We design a caption-based evaluation system that leverages large language models (LLMs) to predict content complexity 
and determine the optimal compression ratio for a given image, taking into account factors critical to human perception.
Trained on images with diverse compression ratios,
\vae demonstrates robust  performance in image reconstruction. We also utilize its variable-length latent representations to train Diffusion Transformers (DiTs) for ImageNet generation. By  optimizing token allocation, \vae improves the FID score over fixed-ratio baselines trained with the same flops and boosts the  inference throughput by 18.5\%. 
\end{abstract}    
\section{Introduction}
\label{sec:intro}

Image tokenizers compress high-resolution images into low-dimensional latent features to generate compact and meaningful representations~\cite{esser2020taming,kingma2022autoencodingvariationalbayes, magvitv2, yu2024an,shen2022dash,nasbench360,fsq}.

Despite their effectiveness, most existing tokenizers use a fixed compression ratio, encoding images into feature vectors of exactly the same dimensions, regardless of their content. 
However, different images contain varying levels of detail,  which suggests that a one-size-fits-all approach to compression may not be optimal. 
Indeed, traditional codecs like JPEG~\cite{jpeg} typically produce different file sizes based on the spatial frequency of the images, even  when set to the same quality level. 

\begin{figure*}
    \centering
\includegraphics[width=\linewidth]{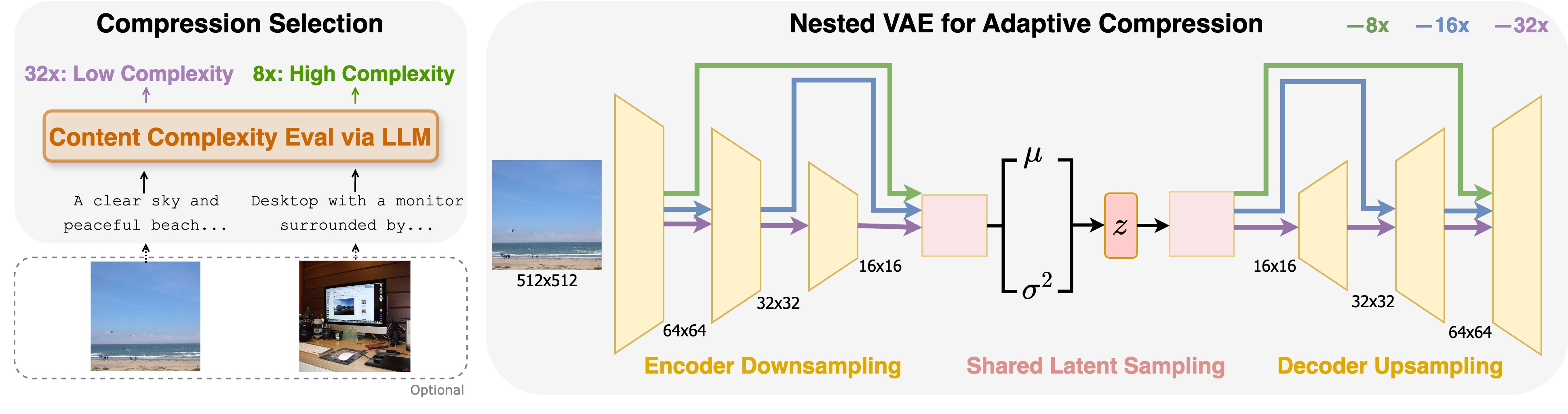}
\caption{\textbf{Content-Adaptive Tokenization.} \vae uses an LLM to evaluate the content complexity and determine the optimal compression ratio based on the image's text description. The image is processed by a nested VAE architecture that dynamically routes the input according to the selected compression ratio.  The resulting latent representations thus have varying spatial dimensions. Images shown in the figure are taken from COCO 2014 \citep{coco}.}
    \label{fig:intro}
\end{figure*}

Moreover, using the same representation capacity for all images can compromise both the quality and the efficiency of the tokenizer. Over-compressing complex images may result in the loss of important visual details, while under-compressing simple images can lead to inefficiencies in training downstream models, as additional compute is wasted on processing redundant information. 
Several recent studies have proposed to adjust the number of tokens used at inference time based on the  compute budget~\cite{yan2024elastic}. 
However, these methods overlook the intrinsic complexity of images when training the tokenizers. Besides, they do not account for the downstream use cases in the tokenizer design.
For example, image tokenizers are often used to produce inputs for latent diffusion models (LDMs)~\citep{rombach2021highresolution}  and perform text-to-image generation, where only the user's text prompt is available at inference time. Nonetheless, existing work all require image inputs to perform adaptive tokenization.

In this work, we present \textbf{C}ontent-\textbf{A}daptive \textbf{T}okenizer (CAT), which dynamically allocates representation capacity based on image complexity to improve both compression  quality and computational efficiency. To achieve this, we propose a text-based image complexity evaluation system that leverages large language models (LLMs) to predict the optimal compression ratio given the image  description. Then, we train a single unified variational autoencoder to generate latent features of variable shapes (Figure~\ref{fig:intro}).

Our complexity evaluation system is designed to accurately reflect the content complexity,
while being compatible with diverse downstream tasks, including text-to-image generation with LDMs.
Specifically, we use the text description of an image to prompt an LLM and generate a complexity score. 
The text description  includes the image caption and answers to a set of perception-focused queries, such as ``\textit{are there human faces/text}", which are designed to help  identify elements     sensitive to human perceptions.
Based on the complexity score, the image is classified into one of 8x, 16x, or 32x compression. A higher ratio means that we can compress a simpler image more   aggressively. 

Then, we develop a nested variational autoencoder (VAE) architecture that can perform multiple levels of compression within a single model. This is achieved by routing the intermediate outputs from the encoder  downsampling blocks to a shared middle block
to generate variable-dimensional Gaussian distributions. From these, we can sample latent features of different spatial resolutions.  

We train the nested VAE on images with diverse complexity, specifically using the compression ratios produced by our LLM evaluator.
We analyze its reconstruction performance on a variety of datasets, including natural scenes (COCO \cite{coco}, ImageNet \cite{imagenet_cvpr09}), human faces (CelebA~\cite{celeba}), and text-heavy images (ChartQA \cite{chartqa}). On complex images featuring human faces or text, 
\vae substantially improves the reconstruction quality, reducing the rFID by 12\% on CelebA and 39\% on ChartQA relative to fixed-ratio baselines. On natural images like ImageNet, \vae maintains the reconstruction quality while using 16\% fewer tokens. 
 
We further validate the effectiveness of CAT in image generation by training Latent Diffusion Transformers (DiTs)~\cite{Peebles2022DiT}. 
Due to its content-adaptive representation, \vae more effectively captures both high-level and low-level information in image datasets  compared to fixed-ratio baselines, hence accelerating the diffusion model learning process.
We demonstrate that \vae achieves an FID of 4.56 on class-conditional ImageNet generation, outperforming all fixed-ratio baselines trained with the same flops. Additionally, \vae improves inference throughput by 18.5\%. 
Beyond the quality and speed improvements, we show that \vae enables controllable generation at various complexity levels, allowing users to specify the number of tokens to represent the images based on practical needs.

To summarize, we introduce \vae, an image tokenizer that enables: 
(1) \textbf{Adaptive Compression:} It compresses images into variable-length latent representations based on content complexity, leveraging  an LLM evaluator and a nested VAE model; (2) \textbf{Faster Generative Learning:} It boosts the efficiency of learning latent generative models by effectively representing both high-level and low-level image information;  (3) \textbf{Controllable Generation:} It enables generation at various complexity levels based on user specifications.
Overall, \vae represents a crucial step towards efficient and effective image modeling, with promising potential for extension to other visual modalities, such as video.
\section{Related Work}
\label{sec:related}
\paragraph{Visual Tokenization.} Existing visual tokenizers use diverse architectures and encoding schemes. Continuous tokenizers map images into a continuous latent space, often utilizing the VAE architecture \cite{kingma2022autoencodingvariationalbayes} to generate Gaussian distributions for sampling latent features. Discrete tokenizers like VQ-VAE \cite{vqvae} and FSQ \cite{fsq} use quantization techniques to convert latent representations into discrete tokens. While our experiments focus on the continuous latent space, the proposed adaptive image encoding method is compatible with both continuous and discrete tokenizers.

\paragraph{Adaptive Compression.} Traditional codecs, such as JPEG~\cite{jpeg} for images and H.264~\cite{h264} for videos, apply varying  levels of compression based on the input media and the desired quality,  resulting in files of different sizes.
In the field of deep learning, a line of work studies adaptive patching for Vision Transformers~\cite{VIT} via patch dropout or merging  \cite{rao2021dynamicvit, yin2022avit,bolya2022tome, tokedropout}. \citet{mixedres}~use mixed-resolution patches to obtain variable-length token sequences. However, these methods are tailored for visual understanding tasks and cannot be used to generate images.

Developing adaptive tokenizers capable of image generation remains underexplored. ElasticTok~\cite{yan2024elastic}, a concurrent work to ours, employs a random masking strategy to drop the tail tokens of an image when training the tokenizer. This allows for using an arbitrary number of tokens to represent an image at inference time. 
However, by assigning random token lengths to training images, ElasticTok overlooks the inherent complexity of the visual content. 
Another concurrent work, ALIT~\cite{duggal2024adaptivelengthimagetokenization}, iteratively distills 2D image tokens into 1D latent tokens to reduce the token count.
Unlike ALIT, \vae compresses images based on   complexity predicted from captions. Our approach enables adaptive allocation of representation capacity using only text descriptions, without directly observing the images.

\paragraph{Multi-Scale Feature Extraction.} A final line of relevant research involves designing neural networks that effectively extract multi-scale features. \vae builds upon VAE and adds skip connections inspired by U-Net~\cite{unet} and Matryoshka representation learning~\cite{kusupati2022matryoshka,cai2024matryoshka,mdm}. In parallel, transformer-based multi-scale feature extractors have also been explored in \citep{Transframer, hu2024matryoshka,shen2023cross, shen2024ups,Roberts2021AutoMLDD,llmtags}. 
We opt for a convolutional tokenizer architecture due to its strong empirical performance.

\section{Method}
In this section, we introduce \vae for adaptive image tokenization. We begin by discussing how to measure and predict image complexity.
Then, we introduce the \vae architecture for performing compression at different ratios.

\begin{figure}[t!]
  \centering
  \begin{subfigure}{0.63\linewidth}
    \includegraphics[width=\linewidth]{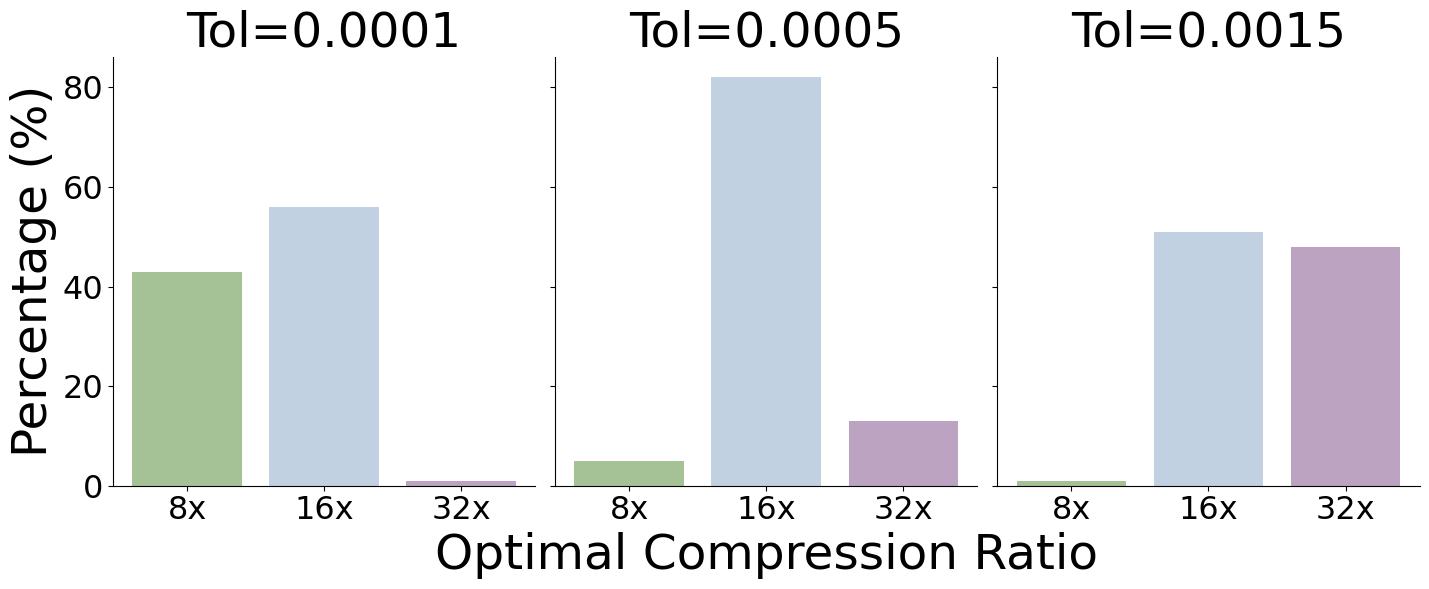}
    \vspace{-1.5cm}
  \end{subfigure}
  \begin{subfigure}{0.3\linewidth}
     \scriptsize
     \begin{tabular}{lc}
    \toprule
    Metric & Pearson $r$ \\
    \midrule
    JPEG & 0.31 \\
    MSE &  0.36\\
    LPIPS & 0.23 \\
    Caption  &\textbf{0.55}\\
    \bottomrule
    
  \end{tabular}
  \end{subfigure}
  \caption{\textbf{Left:} Maximum acceptable compression ratios for COCO images  under different error tolerance. We can compress most images more aggressively without compromising reconstruction quality. \textbf{Right:} Pearson correlation between various metrics and max acceptable compression ratio with tolerance 0.0015.}
   \label{fig:tolerance}
   \label{tab:correlation}
\end{figure}

\renewcommand{\thefootnote}{\arabic{footnote}}
\subsection{Proof of Concept}

\subsubsection{How Much Can We Actually Compress?}
\label{sec:proof:motivation}

A key question in this work is to determine how much an image can be compressed without significant loss of quality. 
To explore this, we analyze the reconstruction performance of existing  tokenizers with various compression ratios. 
We take the open-source image tokenizers from LDM\footnote{LDM released a series of VAE tokenizers with diverse compression ratios and trained in a controlled setting. Most other tokenizers, such as \texttt{stabilityai/sd-vae-ft-mse}, only have one compressed ratio.}~\cite{rombach2021highresolution} with 8x, 16x and 32x compression ratios and compute their reconstruction mean squared error (MSE) on 41K 512$\times$512 images from the COCO 2014 test set \cite{coco}.  Our analysis reveals that for 28.3\% of the images, 32x compression  results in less than a 0.001 MSE increase  compared to 8x compression, while reducing the token count by a factor of 16. We also compute the best MSE among all compression ratios for each image and determine the  maximum acceptable compression ratio under a tolerance $\tau$. That is, denote the compression ratio as $f$, we want to find
\begin{align}
   \argmax_{f\in\{8, 16, 32\}} \big( MSE_{f} -\min_{f'\in\{8, 16, 32\}} MSE_{f'} \big) < \tau.
\end{align}
Figure~\ref{fig:tolerance} shows that 56\% of the images can be compressed at least to 16x with negligible (0.0001) increase in MSE\footnote{Note that the average MSE across all images for 8x LDM VAE is 0.0039, so a 0.0001 tolerance should be acceptable.}. A large portion of natural images can be compressed more aggressively while maintaining the same quality level as a fixed 8x tokenizer.

On the other hand, our visual inspection reveals that images with fine-grained elements like text have much better reconstruction quality at 8x compression compared to 32x (see for example row 3 and 4 in Figure~\ref{fig:example}). This suggests that more tokens are required to accurately reconstruct low-level details in such images.
The above results 
provide strong motivation for developing a tokenizer with adaptive compression ratios. Accordingly, we set the target   ratios for \vae to be 8, 16, and 32. 

\subsubsection{Limitations of Existing Complexity Metrics}
\label{sec:proof:limitation}

\begin{figure}[t!]
    \centering
    \vspace{-2mm}
    \includegraphics[width=\linewidth]{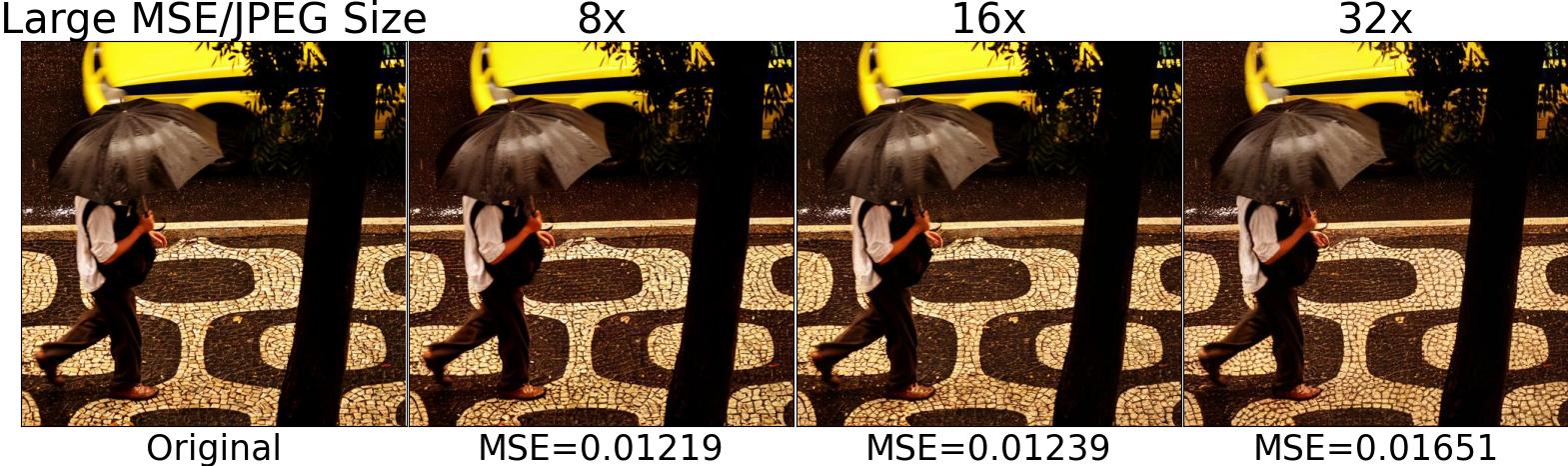}
    \vspace{2mm}
\includegraphics[width=0.99\linewidth,right]{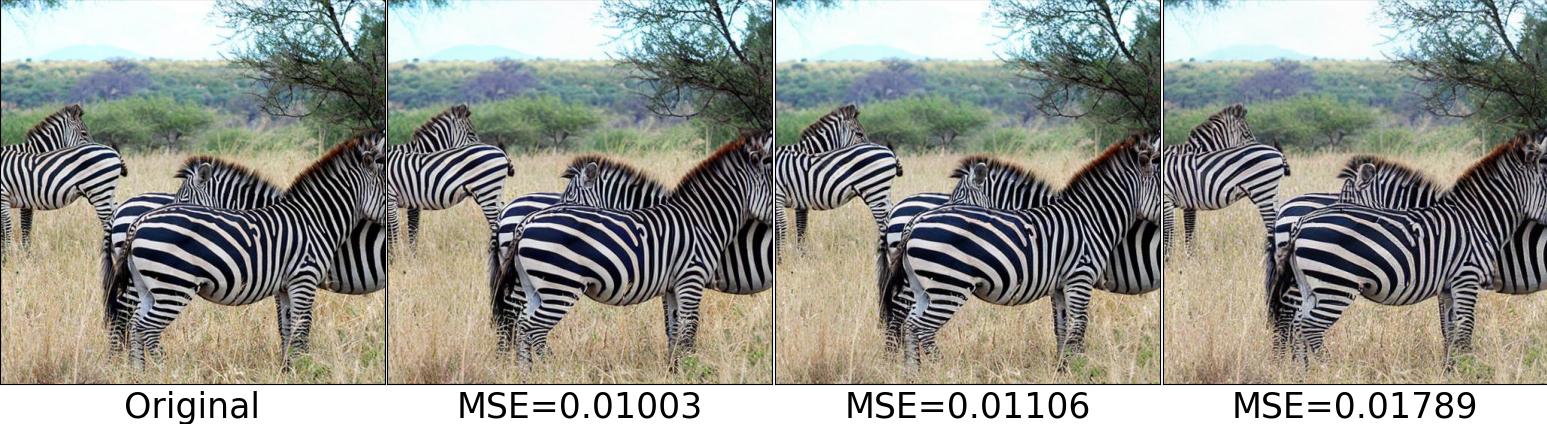}
    \includegraphics[width=\linewidth]{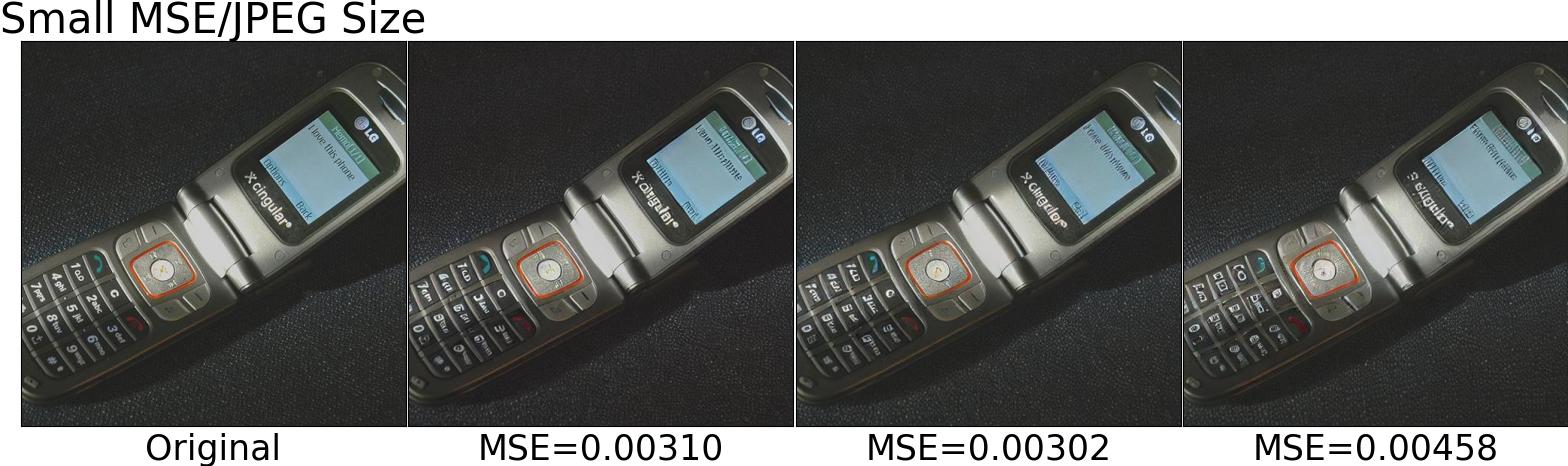}
\includegraphics[width=0.99\linewidth,right]{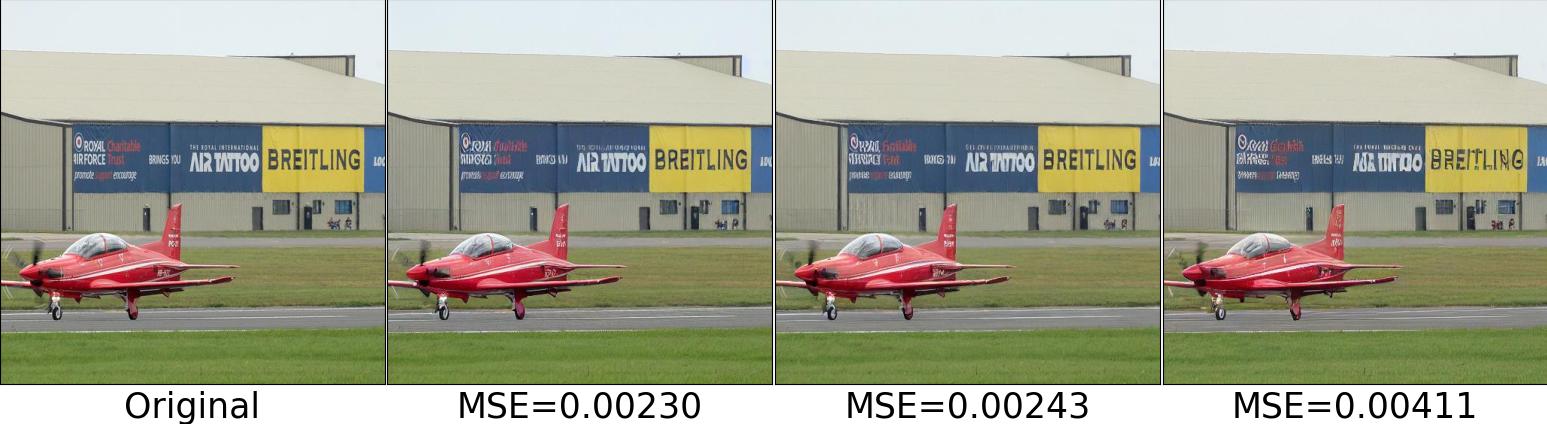}
    \caption{\textbf{Existing metrics can misjudge image complexity.} Metrics like JPEG size,  MSE, and LPIPS consider images with high contrast and repetitive patterns as complex but underestimate the complexity of text-heavy images that are  more challenging for human perception (note the distortion in the bottom two rows).  Images shown in the figure are taken from COCO 2014 \citep{coco}. }
    \label{fig:example}
    \vspace{-3mm}
\end{figure}

Next, we want to identify a metric for predicting the optimal compression ratio given an image. We explore some existing options,  categorized into two groups: (1) metrics produced by traditional codecs, i.e., the JPEG file size; (2) metrics based on pretrained VAEs\footnote{We use \texttt{stabilityai/sd-vae-ft-mse} for this analysis.}, including reconstruction MSE and LPIPS \cite{zhang2018perceptual}, which measures the L2 distance of VGG Net \cite{vgg}  activations  between the original and reconstructed images.
We first compute these metrics on the COCO dataset and analyze their correlation with the maximum acceptable compression ratio under $\tau=0.0015$. However, Table~\ref{tab:correlation} shows that  the Pearson correlations are  relatively low.

After that, we manually inspect the images with large JPEG sizes and MSEs. We note that images featuring repetitive patterns, such as grass, forests, and animals  like giraffes and zebras consistently  show high complexity  metrics. Indeed, JPEG compression can be inefficient for images with sharp edges and high contrast. A  single-pixel shift in a zebra image can toggle pixel values between black and white, significantly increasing the pixel-wise MSE.
However, as the top rows in Figure~\ref{fig:example} show, large MSEs do not always notably affect visual quality. 
For example, we can easily recognize the zebra and may not perceive the differences resulting from various compression ratios.

On the contrary, we find that many images with low considered metrics in fact have low fidelity. These images often contain visual elements like human faces or text, where even slight distortions can degrade visual quality (Figure~\ref{fig:example}, bottom rows). Despite this, these images have low MSEs possibly because the critical elements occupy only small portions of the images. Thus,  metrics like JPEG size, MSE, and LPIPS might not effectively capture  details crucial to human perception. Contrary to the predicted complexity,  we actually want to use a small compression ratio for text-heavy images, and a large compression ratio for the zebra.

Lastly, the considered metrics all require images as input and cannot be used to measure  complexity for text-to-image generation tasks, where no image is  available at inference time.
Given all these limitations of existing metrics, 
we seek a new method that is  independent of pixel data and aligns with human perception to predict image complexity.

\subsection{Complexity Evaluation via Captions and LLMs}

Image generation typically involves users providing a prompt that describes the desired image content. To better align with such real-world use cases, we leverage the text description of an image to measure its content complexity. 
 
We propose a three-stage complexity evaluation system: (1) obtaining the text description, (2) prompting an LLM to output a complexity score, and (3) classifying the  score into a compression ratio. The text description consists of both the image caption and responses to a pre-defined set of perception-focused questions ``\textit{Are there [obj]?}" where $\textit{obj}\in\{\textit{human faces}, \textit{text}\}$. This set can be expanded to accommodate different needs. When images are available, we use InstructBlip \cite{InstructBLIP} to generate  the caption and the responses. Otherwise, users need to provide the required  description in text.

In stage 2, the text description is processed by an LLM to assess complexity. We use Llama 3 70B Instruct~\cite{dubey2024llama3herdmodels} in this work. To ensure consistency in scoring, we design a detailed prompt consisting of the evaluation instructions; the output scale, i.e., an integer score ranging from 1 to 9, where higher scores indicate greater complexity; important factors for scoring, such as semantic complexity (objects, scenes), visual complexity (color, lighting, texture), and perceptual complexity (presence of faces and text); and lastly, specific examples for each score as demonstrations. We provide the prompt we use in Appendix \ref{sec:appen:prompt}.

We divide the scores into three intervals: $[1, a]$, $(a, b]$, and $(b, 9]$, where $a < b\in \mathbb{Z}^+$. After obtaining the score from the LLM,  we classify it into one of 8x, 16x, and 32x compression ratios, with higher complexity scores corresponding to lower compression ratios.
The threshold points $a$ and $b$ are selected to achieve an average compression ratio of approximately 16x across all training data, allowing us to make a fair comparison with fixed 16x baselines.

Formally, denote the training distribution as $\mathcal{X}$, input resolution as $r$, the compression ratio of an image $x\in \mathcal{X}$ as $f(x)\in \{f_1, f_2, f_3|f_1=8,f_2=16,f_3=32\}$, and the target average compression ratio as $\bar f:=16$. After collecting the complexity scores for all training images, we set $a,b$ to meet the target compression ratio:
{\small
\begin{align}
    \mathbb{E}_{x\in\mathcal{X}}[\frac{r^2}{f(x)^2}] \approx \sum_{x\in \mathcal{D}} p\big(f(x)\big) \frac{r^2}{f(x)^2} \approx \frac{r^2}{\Bar{f}^2}
\end{align}}
There could be multiple sets of thresholds that achieve the target compression ratio. We show in Section~\ref{sec:eval_recon:ablation} that a more diverse distribution of compression ratios leads to better empirical performance. We discuss the exact training data we use and the threshold selection in Section~\ref{sec:eval_recon:setup}.

Finally, we  verify  the proposed caption complexity indeed provides a good estimation of the optimal compression ratio. We compute the correlation between our complexity score and the maximum acceptable compression ratio for COCO images and find that it surpasses all existing metrics (Table~\ref{tab:correlation}). Meanwhile, the compression ratio selected by our caption score achieves an exact  agreement of 62.39\% with the maximum acceptable compression ratio. We also manually inspect the images and confirm that perceptually challenging images are assigned high caption complexity.

\subsection{Nested VAE for Adaptive Compression}
To reduce training and storage costs,  we  introduce a nested structure to the standard VAE architecture~\cite{kingma2022autoencodingvariationalbayes}
to enable multiple compression ratios within a single model. In the  standard VAE architecture, the encoder consists of multiple downsampling blocks followed by an attention-based middle block. The decoder consists of an attention-based middle block followed by upsampling blocks. This symmetrical design is reminiscent of U-Net \cite{unet} and Matryoshka networks \cite{kusupati2022matryoshka} for multi-scale feature extraction. Inspired by these works, we leverage the intermediate outputs of the downsampling blocks to enable adaptive compression. We describe the proposed architecture below. See Figure~\ref{fig:intro} for illustration.

\paragraph{Skip Connection with Channel Matching.} 
Denote the feature shape under the largest compression ratio as $(c_3, \frac{r}{f_3}, \frac{r}{f_3})$, where  $c_3$ is the channel dimension.
We observe that, in the standard VAE encoder, the spatial dimension of the intermediate outputs from the downsampling blocks  decreases by a factor of 2 with each additional block.  
This means that the output of the second-to-last downsampling block naturally has shape $(c_2, \frac{r}{f_2}, \frac{r}{f_2})$, and the output of the third-to-last downsampling block has shape $(c_1, \frac{r}{f_1}, \frac{r}{f_1})$. 
An immediate thought is to directly route these intermediate outputs to the middle block to generate latent features. 
However, since the channel dimensions of these intermediate outputs vary, we leverage ResNet blocks \cite{resnet} for channel matching. Let the latent channel dimension of the VAE be $c$. Applying channel matching enables us to transform intermediate features of shape $(c_{n}, \frac{r}{f_n}, \frac{r}{f_n})$ to $(c, \frac{r}{f_n}, \frac{r}{f_n})$ for $n=1,2,3$. This will be the shape of the latent parameters.

For the decoder, similarly,
we  add skip connection with channel matching and route the output from the decoder's middle block to the corresponding upsampling block.
For the compression ratio $f_n$, we bypass the first $n - 1$ upsampling blocks  to ensure the decoder output has the same resolution as the original image. 

\paragraph{Shared mean/variance parametrization.} In the encoder, features after channel matching are directed to the middle block to generate the parameters of the latent distribution. For the \vae architecture, we share the middle block for all compression ratios to maintain scale consistency of the parameterized mean and variance. The convolutional design of the middle block allows it to process inputs of varying spatial dimensions, as long as the channel dimension is aligned. Thus, for all $n\in\{1,2,3\}$, the mean $\mu_n$, variance $\sigma^2_n$, and sample $z_n$ of the Gaussian distribution all have shape $(c, \frac{r}{f_n}, \frac{r}{f_n})$, which is the original input compressed at ratio $f_n$.

\paragraph{Increasing parameter allocation for shared modules.} 
Images assigned smaller compression ratios do not go through the later downsampling blocks and are directed straight to the middle block. 
The middle block is thus tasked with handling multi-scale features. To improve its capacity, we allocate more parameters to the middle block by increasing the number of attention layers. 

\begin{table*}[htbp]
\resizebox{\textwidth}{!}{
\centering
\Large
\begin{tabular}{lllcccccccccccc}
\toprule
\multirow{2}{5em}{Average Compression}   &&&& COCO & & &ImageNet& && CelebA &&& ChartQA\\
  \cmidrule(lr){4-6} \cmidrule(lr){7-9} \cmidrule(lr){10-12} \cmidrule(lr){13-15}
   & &&rFID$\downarrow$ &LPIPS$\downarrow$&  PSNR $\uparrow$&  rFID$\downarrow$ & LPIPS$\downarrow$& PSNR $\uparrow$& rFID$\downarrow$ & LPIPS$\downarrow$& PSNR $\uparrow$& rFID$\downarrow$ & LPIPS$\downarrow$& PSNR $\uparrow$ \\
\toprule
8& Fixed &8x  &0.48&0.10&30.95&0.24&0.095&33.86&1.86&0.028&45.36&8.21&0.019&36.98\\
\midrule
\multirow{3}{5em}{16}&Fixed &16x&0.66&0.16&29.79&\textbf{0.38}&\textbf{0.15}&30.45&2.25&0.059&41.84&8.67&0.029&33.48 \\
&Adaptive &JPEG&0.72&0.17&30.11&0.51&0.16&30.61&6.57&0.14&36.47&10.17&0.048&31.54\\
&Adaptive & \vae \textbf{(Ours)}&\textbf{0.65}&\textbf{0.15}&\textbf{30.19}&0.46&\textbf{0.15}&\textbf{30.62}&\textbf{1.97}&\textbf{0.051}&\textbf{42.43}&\textbf{5.27}&\textbf{0.021}&\textbf{36.45}\\

\midrule
32&Fixed & 32x &1.18&0.26&26.93&0.81&0.25&27.48&6.10&0.16&36.35&10.79&0.045&30.99 \\
\bottomrule
\end{tabular}}
 \caption{\textbf{Reconstruction results.} All models have latent channel $c=16$. \vae outperforms fixed 16x and JPEG baselines on most metrics.}
\label{tab:reconstruction}
\end{table*}

\paragraph{Training.}
While existing adaptive tokenizers like ElasticTok \cite{yan2024elastic} do not consider the different complexity levels within the training data, we explicitly incorporate content complexity into the training process to learn feature extraction at different granularity. For each training example, we first obtain the compression ratio from the LLM evaluation system. Then, the image is processed only by the layers dedicated to its  compression ratio.

Similar to prior works \cite{kingma2022autoencodingvariationalbayes,esser2020taming}, we use a joint objective that minimizes reconstruction error, Kullback-Leibler (KL) divergence, and perceptual loss. Specifically, we use ${L}_1$ loss for pixel-wise reconstruction. To encourage the encoder output $z$ towards a normal distribution, KL-regularization  is added: $\mathcal{L}_{\text{KL}}(z) := \mathbb{KL}(q_\theta(z|x) \| p(z)\big)$, where $\theta$ is the encoder parameters and $p(z)\sim \mathcal{N}(0, \mathbf{I})$. The perceptual loss consists of the LPIPS similarity \cite{zhang2018perceptual} and  a loss based on the internal
features of the MoCo v2 model \cite{moco}. Beyond these, we   train our tokenizer in an adversarial manner \cite{gan} using a patch-based discriminator $\psi$. This leads to an additional GAN loss $\mathcal{L}_{\text{GAN}}(x, \hat x, \psi)$. Thus, our overall objective is:
{
\begin{align}
\mathcal{L} =  \min_{\theta} \max_\psi  \;&\mathbb{E}_{x\in\mathcal{X}}\Big[\mathcal{L}_{\text{rec}}(x, \hat x) + \beta\mathcal{L}_{\text{KL}}(z)\nonumber \\& + \gamma\mathcal{L}_{\text{perc}}(\hat x)+ \delta\mathcal{L}_{\text{GAN}}(x, \hat x, \psi)\Big],
  \label{eq:loss}
\end{align}}
where $\beta,\gamma,\delta$  are the weights for each loss term. 
To simplify implementation, we first sample a compression ratio for each GPU and ensure a batch of training data contains images with the same compression ratio.
However, different GPUs can have different compression ratios.

\section{Image Reconstruction}
\label{sec:eval_recon}

We first evaluate \vae on image reconstruction. We will present downstream generation results in Section~\ref{sec:eval_gen}.

\subsection{Setup}
\label{sec:eval_recon:setup}

\paragraph{Model and Training.} 
We use a nested VAE architecture with  six downsampling blocks; the output channels are 64, 128, 256, 256, 512, 512. We use 8 attention layers for the middle block. 
The  latent channel $c$ is  16 for experiments in Table~\ref{tab:reconstruction}, but we study its effect as an ablation study in Table~\ref{tab:latentdim}. The total number of parameters is 187M. 

For training data, we use a collection of 380M licensed Shutterstock images with input resolution 512.
After obtaining the complexity scores, we find that two sets of threshold points, $(a, b)\in\{(4, 7), (2, 8)\}$,
both achieve an average compression ratio of approximately 16x. However, since $(4, 7)$ leads to a more diverse distribution and better emprical performance (see Table~\ref{tab:train_dist} and ablation studies in Section~\ref{sec:eval_recon:ablation}), we use it in the final setup of \vae. All models including the baselines are trained using a global batch size of 512
on 64 NVIDIA A100 GPUs for 1M steps. Further architecture and training details (e.g., loss weights, optimizer, and learning rate schedule) can be found in Appendix \ref{sec:appen:recon}.

\paragraph{Baselines.} We compare \vae against fixed compression ratio baselines that use the same VAE architecture but without the nested structure.
To study the effect of caption-based complexity, we train another nested VAE using the JPEG file size of the image  as the complexity metric. We ensure all  models have average 16x compression. See Appendix~\ref{sec:appen:recon:baseline} for more baseline details.

\paragraph{Evaluation Datasets and Metrics.} We evaluate the reconstruction performance on four datasets: COCO~\cite{coco} and ImageNet \cite{imagenet_cvpr09}, representing natural images; CelebA \cite{celeba} and ChartQA \cite{chartqa}, representing perceptually challenging images.  We report reconstruction FID (rFID),   LPIPS, and PSNR \cite{psnr} as the performance metrics. 

\begin{table}[t!]
\centering
\Large
\resizebox{0.8\linewidth}{!}{
\begin{tabular}{@{}llcccc@{}}
    \toprule
   \multirow{2}{3em}{Eval Dataset} & \multirow{2}{6em}{Compression Method} &\multicolumn{3}{c}{Eval  Distribution}&\multirow{2}{5em}{Latent Dim}\\
    \cmidrule(lr){3-5}
&   & 8x & 16x & 32x &\\
    \midrule
\multirow{2}{4em}{COCO}&\vae&9\%&54\%&37\%&31.87\\
&JPEG&10\%&54\%&36\%&32.43\\
\midrule
\multirow{2}{4em}{ImageNet}&\vae&6\% &49\% & 45\%	&29.32\\
&JPEG&9\%&	49\%&42\%&31.24\\
\midrule
\multirow{2}{4em}{CelebA}&\vae&17\%&	83\%&	0\%&39.29\\
&JPEG&0\%	&0\%&	100\%&16\\
\midrule
\multirow{2}{4em}{ChartQA}&\vae&96\%&	4\%&	0\%&63.02\\
&JPEG&0\%&	3\%&97\%&16.61\\
    \bottomrule
  \end{tabular}}
  \caption{\textbf{Test data distribution and average spatial dimension ($\frac{r}{f}$) of the latent features.}  The numbers  denote the proportion of images for each dataset. Compared to fixed 16x baseline, which has a latent dimension of $\frac{512}{16}=32$, \vae uses smaller latents for natural images and larger latents for CelebA and ChartQA. }
  \label{tab:test_dist}
\end{table}

\subsection{Main Results}
\begin{figure*}
\vspace{-3mm}
    \centering
    \includegraphics[width=0.95\linewidth]{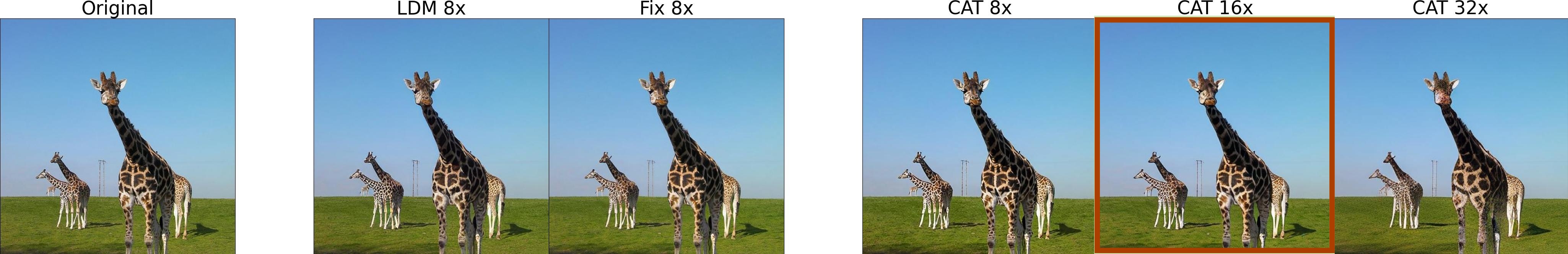}
    \includegraphics[width=0.95\linewidth]{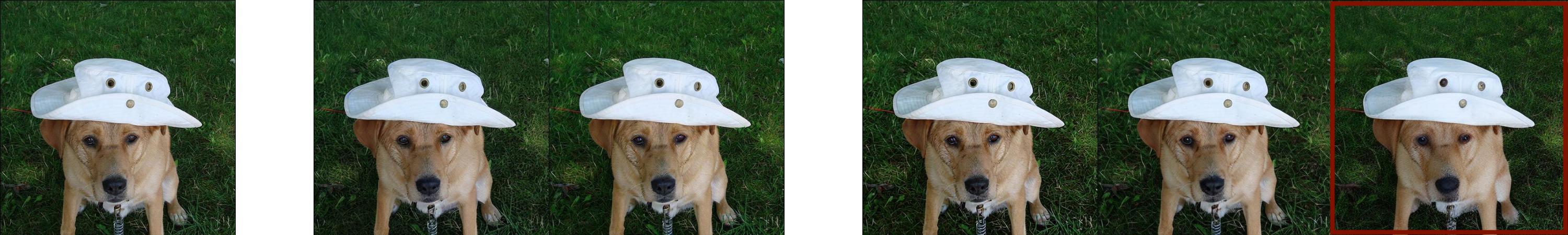}
    \includegraphics[width=0.95\linewidth]{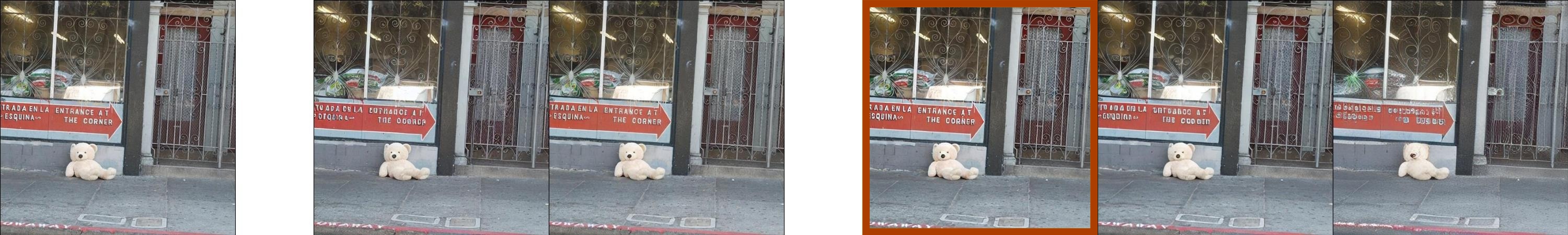}
    \includegraphics[width=0.95\linewidth]{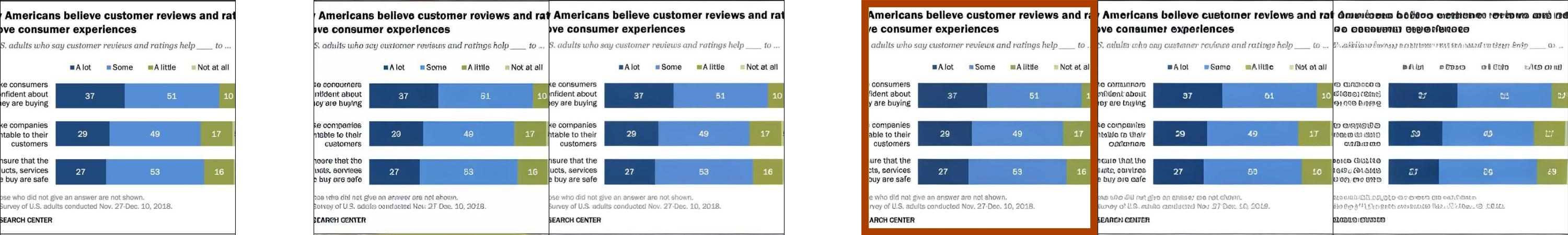}
    \includegraphics[width=0.95\linewidth]{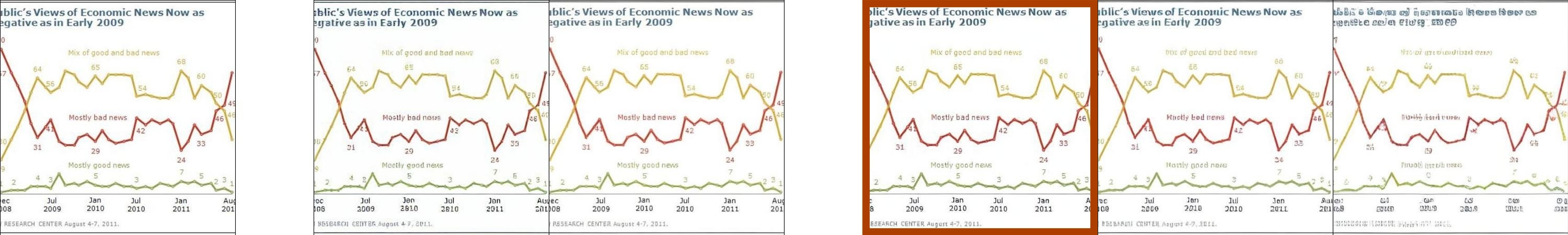}
    \caption{\textbf{We highlight the compression ratio selected by our proposed caption complexity in red.} On simpler images (top two rows), adjusting the \vae compression ratio   does not significantly affect quality. On more complex images (bottom three rows), the impact is substantial.  Also note that \vae's text reconstruction  is comparable with fixed 8x baseline and better than pretrained LDM VAE.  Images shown in the figure are taken from COCO 2014 \citep{coco} and ChartQA \citep{chartqa}.}
    \label{fig:reconvisual}
\end{figure*}
Table \ref{tab:reconstruction} presents the image reconstruction results of \vae and various baselines. For fixed compression methods, the 8x compression ratio achieves substantially better performance than the 16x and 32x compression ratios, which shows  that reducing the compression ratio is an effective strategy to improve reconstruction at the cost of increased computational expense. Then, we compare our method with the fixed 16x  baseline. On COCO and ImageNet, \vae generally outperforms the baseline, with only a slight drop in rFID on ImageNet. 
However, the average dimension of \vae latent features is 31.87 for COCO and 29.32  for ImageNet, both of which are smaller than the baseline dimension of 32~(Table~\ref{tab:test_dist}). 
This shows \vae can effectively learn compact representations  for natural images.
On CelebA and ChartQA, \vae\ significantly outperforms the baselines on all metrics. On ChartQA, \vae\ even surpasses the fixed 8x baseline,  proving its efficacy in capturing visual details.

We also compare \vae with training the same adaptive architecture but using JPEG size as the complexity metric. Across all datasets, \vae achieves better rFID, LPIPS, and PSNR. While we ensure both tokenizers have the same \textit{training} compression ratio distribution, the compression ratio distribution of the evaluation datasets varies significantly (Table~\ref{tab:test_dist}). Notably, since JPEG size often cannot capture perceptually important factors (see discussion in Section \ref{sec:proof:limitation}), nearly all images in CelebA and ChartQA are assigned the highest 32x compression ratio. Thus, \vae significantly outperforms JPEG on these two datasets, showing the effectiveness of caption-based metric and LLM evaluation in determining an image's intrinsic complexity.

\begin{table}[t!]
\centering
\Large
\resizebox{\linewidth}{!}{
  \begin{tabular}{lcccccccc}
    \toprule
   \multirow{2}{2em}{$(a,b)$} &\multicolumn{4}{c}{Training   Distribution }&
    \multicolumn{4}{c}
    {Reconstruction FID $\downarrow$}\\
    \cmidrule(lr){2-5}\cmidrule(lr){6-9}
& 8x & 16x & 32x  & Average &COCO&ImageNet&CelebA&ChartQA\\
    \toprule
 $(4, 7)$  &10\%&	48\%&42\%	 &	16.0x&\textbf{0.65}&{0.46}&\textbf{1.97}&\textbf{5.27}\\
 $(2, 8)$& 0.5\%&89.5\%&10\%	 &	16.5x&0.67&\textbf{0.43}& 2.58&7.70\\
    \bottomrule
  \end{tabular}}
  \caption{\textbf{Compression ratio distribution affects learning outcomes.} Both settings have an average compression of $\sim$16x, but $(4,7)$ leads to better distribution diversity and empirical results.}
  \label{tab:train_dist}
\end{table}
Figure \ref{fig:reconvisual} shows qualitative examples of progressive reconstruction quality using the learned \vae VAE  as we manually reduce the compression ratio  and use more tokens to represent each image. We highlight the compression ratio selected by our caption metric in red. Different visual inputs have different optimal compression ratios. Natural images with fewer objects and simpler patterns can be accurately reconstructed at 32x, whereas complex images with visual details require lower compression. Thus, the caption-based \vae reconstruction has comparable quality to the fixed 16x baseline on natural images but surpasses it on text-heavy images.
These results further demonstrate the effectiveness of \vae. We include more visualization and comparison with LDM VAEs in Appendix~\ref{sec:appen:gen:visual}.

\subsection{Ablation Studies}
\label{sec:eval_recon:ablation}

We explore several design choices for our tokenizer. First, we study how the distribution of compression ratios affects overall reconstruction. To achieve an average compression ratio of 16, we consider setting the thresholds $(a, b)$ to either $(4, 7)$ or $(2, 8)$. As shown in Table~\ref{tab:train_dist}, the configuration $(4, 7)$ yields a more diverse distribution, whereas $(2, 8)$ results in a distribution that is more concentrated and similar to a fixed 16x tokenizer—making it a less interesting setting. Table \ref{tab:train_dist} also compares the reconstruction performance of these configurations. The thresholds $(4, 7)$ produce better reconstruction metrics across all datasets, possibly because the diversity in compression ratios ensures that all parts of the model are fully trained. Consequently, we adopt $(4, 7)$ as the thresholds for \vae.

\begin{table*}[t!]
\renewcommand\arraystretch{1.2}
\centering
\scriptsize

\resizebox{0.75\linewidth}{!}{
\centering
\begin{tabular}{llcccccc}
    \toprule
   \multicolumn{2}{l}{{DiT-XL/2+Tokenizer}} &FID$\downarrow$   & sFID$\downarrow$  & IS$\uparrow$     & Precision$\uparrow$ & Recall$\uparrow$   & Eval rFLOPs$\downarrow$\\
    \bottomrule
    \multirow{2}{2em}{Fixed}&{LDM VAE}  &10.03&16.88&114.84&0.65&\textbf{0.50}& 1$\times$ \\
    &Fixed 16x&4.78&11.81&187.47&0.72&0.49 & 1$\times$ \\
\midrule
    Adaptive&\vae &\textbf{4.56}&\textbf{10.55}&\textbf{191.09}&\textbf{0.75}&0.49 &\textbf{0.82$\times$} \\
    \bottomrule
 \end{tabular}}
    \caption{\textbf{512$\times$512 class-conditional ImageNet generation results after 400K training steps (cfg=1.5).}  
    All tokenizers have average compression ratio $\bar f=16$ and latent channel $c=16$.  ``rFLOPs" means relative FLOPs.}
    \label{tab:gen}
\end{table*}

\begin{table}[t!]
\centering
\resizebox{0.8\linewidth}{!}{
\begin{tabular}{lccccc}
\toprule
 rFID$\downarrow$&$c$ & {COCO}  &{ImageNet}&  {CelebA} & {ChartQA}\\
\toprule
 & 4&1.25&1.32&5.89&9.45\\
Fixed 16x & 8&1.10&0.61&4.99&\textbf{8.19}\\
 &16&\textbf{0.66}&\textbf{0.38}&\textbf{2.25}&8.67\\
 \midrule
 & 4&1.66&1.10&5.83&9.13\\
 \vae   &8&1.03&0.60&4.54&7.95\\
& 16&\textbf{0.65}&\textbf{0.46}&\textbf{1.97}&\textbf{5.27}\\
\bottomrule
\end{tabular}}
\caption{\textbf{Larger latent channel  $c$ generally improves rFID.}}
\label{tab:latentdim}
\end{table}
We also vary the latent channel dimension $c$ to study its effect on tokenizer performance. As shown in Table  \ref{tab:latentdim}, a larger $c$ leads to better reconstruction metrics. However, consistent with previous studies~\cite{rombach2021highresolution, emu}, we observe a reconstruction-generation trade-off: while increasing $c$ improves reconstruction quality of the tokenizer, it does not necessarily result in better second-stage generative performance. We elaborate on this trade-off in the next section.

\section{Image Generation}
\label{sec:eval_gen}

In this section, we use \vae to develop image generation models for ImageNet dataset. Given the continuous and adaptive nature of \vae, we use the diffusion transformer (DiT) \cite{Peebles2022DiT} as the second-stage model, which is capable of handling variable-length token sequences. DiT takes the noised latent features as input, applies patching to further downsample the input, and  uses a transformer architecture to predict the added noise.

\subsection{Setup}
Following \citet{Peebles2022DiT}, we utilize DiT-XL with 431M parameters and a patch size of 2. We work with images of input resolution 512. With a 16x compression during tokenization and an additional 2x compression during patching, the number of patches (referred to as ``tokens" hereafter) representing each image is  $(\frac{512}{16\cdot2})^2=256$.

Since the ImageNet dataset does not naturally include text captions, we employ InstructBlip to generate captions for the images individually during training. For inference, we use the caption ``\textit{this is an image of [label]}". We follow our scoring system to determine the target number of tokens to generate—specifically, 64 for 32x decoder, 256 for 16x decoder, and 1024 for 8x decoder.

As for baselines, we consider DiT-XL paired with the open-source 16x LDM  VAE and the fixed 16x tokenizer trained in the previous section. We train all models with the same global token batch size of 262,144, which is equivalent to 1,024 images at a 16x compression ratio, and for 400,000 steps on 16 NVIDIA H100 GPUs.
Following  the original DiT work,
we report FID \cite{fid}, Sliding FID \cite{sfid}, Inception Score \cite{is}, precision and recall~\cite{precision} on 50K images generated with 250 DDPM sampling steps and classifier-free guidance \cite{cfg}.
See Appendix \ref{sec:appen:gen} for details.

\subsection{Results}

Table \ref{tab:gen} summarizes the results, showing that \vae\ achieves the best  FID, sFID, IS, and precision among all baselines trained with the same computational resources. We attribute this strong performance to two factors. First, adaptively allocating representation capacity enables more effective modeling of complex images while reducing noise in simpler ones. Second, using fewer tokens for simpler images improves processing efficiency, allowing for more extensive and diverse training within the same computational budget. Specifically, since ImageNet primarily consists of natural images, only a few classes featuring people or fine-grained text receive high complexity scores. On the training dataset, the average token count per image for DiT-\vae\ is 197.44, which is 23\% lower than the 256 tokens used by DiT with fixed 16x tokenizers. During inference, this average increases to 216, leading to an 18.5\% increase in inference throughput (samples per second).

\begin{table}[t!]
\centering
\resizebox{0.65\linewidth}{!}{
\begin{tabular}{lccc}
\toprule
&\vae 8x&\vae 16x&\vae 32x\\
\midrule
FID-50K&4.12&5.02&5.83\\
\bottomrule
\end{tabular}}
\caption{\textbf{We manually adjust the inference token count for \vae with $c=8$ to control the complexity of the generated images.} }
\label{tab:control}
\end{table}

\begin{figure}
    \centering
    \includegraphics[width=\linewidth]{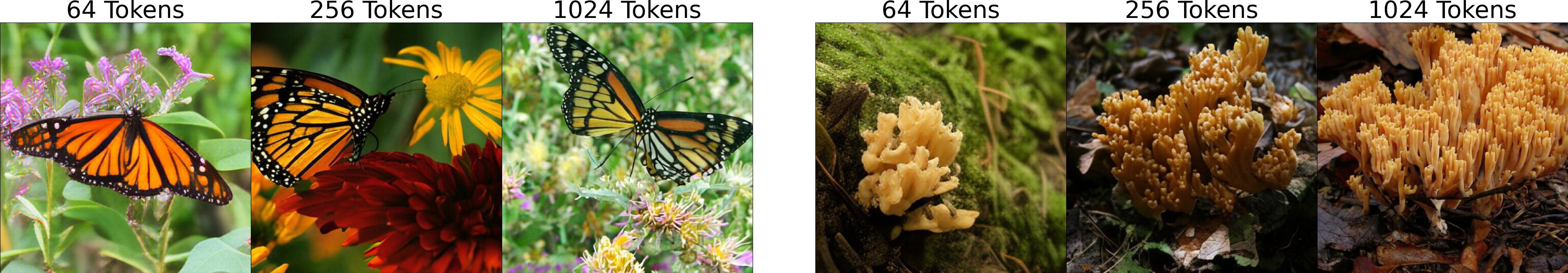}
    \includegraphics[width=\linewidth]{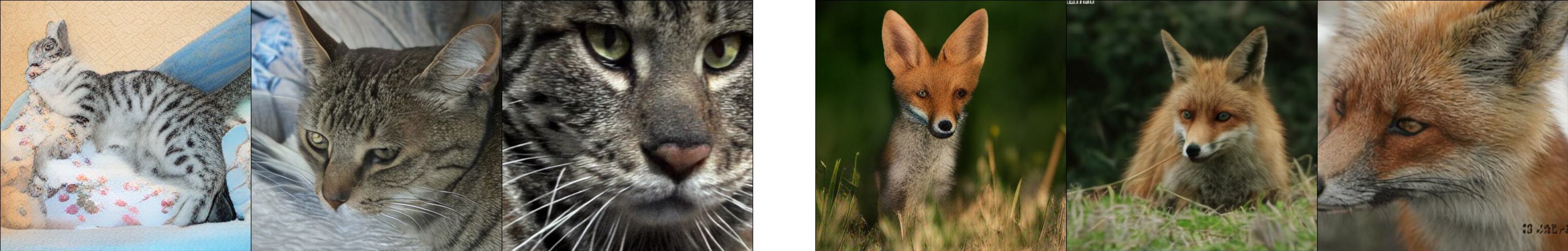}
    \includegraphics[width=\linewidth]{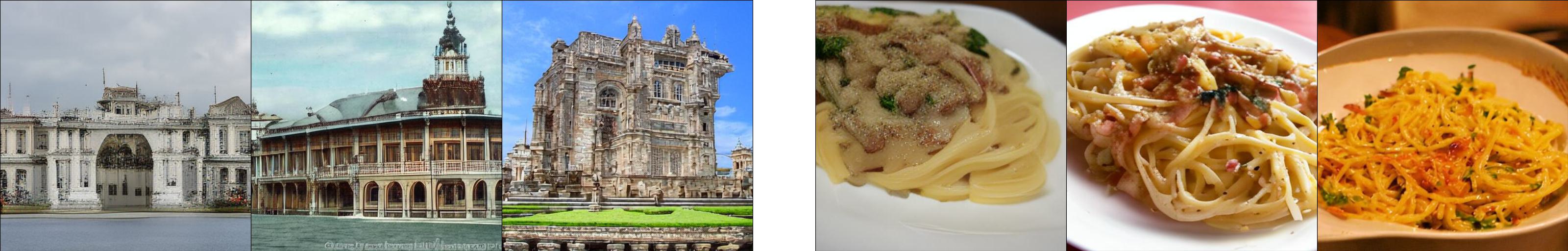}
    \includegraphics[width=\linewidth]{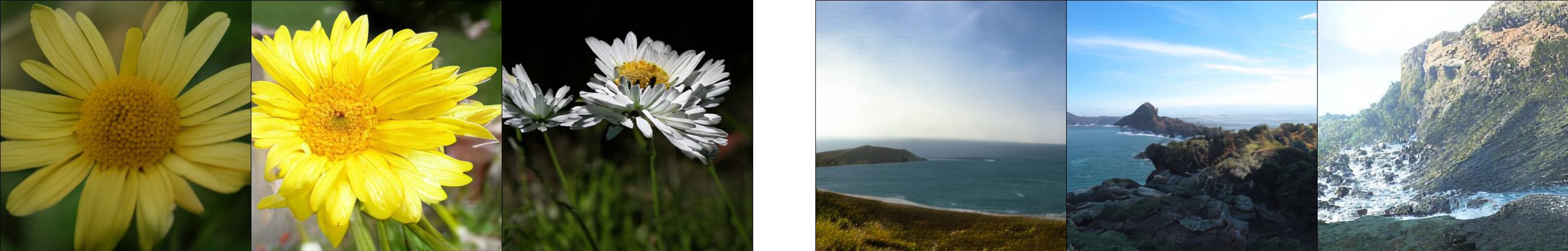}
    \includegraphics[width=\linewidth]{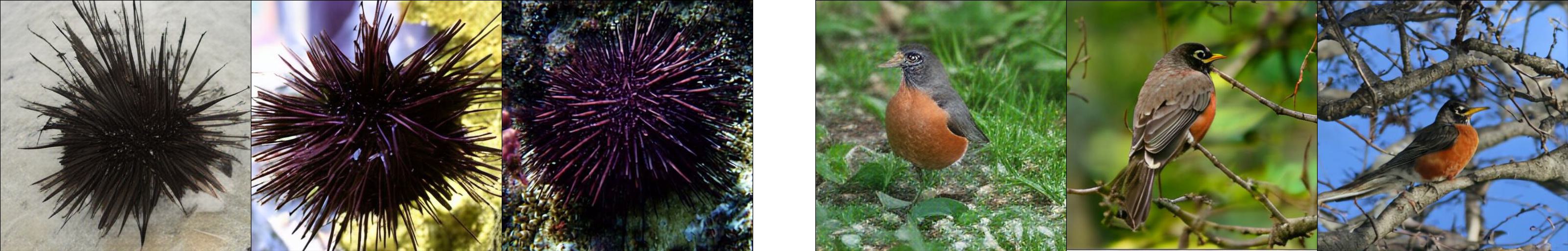}
    \caption{\textbf{Increasing token count (left$\rightarrow$right) for \vae leads to better image quality and higher complexity.}}
    \label{fig:genvisual}
\end{figure}

We study the effect of  manually increasing the number of tokens for DiT-\vae during generation. Table~\ref{tab:control} shows that FID score is significantly improved when using more tokens during image generation.  
We further provide qualitative examples. 
As Figure \ref{fig:genvisual} shows, utilizing more tokens leads to more complex images, such as featuring more objects and more intricate texture. This highlights a side benefit of adaptive tokenization: it enables complexity-controllable generation at no additional training cost.

Lastly, recall that we trained tokenizers with different latent channels in Section \ref{sec:eval_recon:ablation}. Table \ref{tab:latentdimgen} shows the generation performance. While larger $c$ is better for reconstruction, it is not the case for generation. In fact, $c=8$ leads to better average results for both fixed and adaptive settings, and \vae with $c=8$ obtains the best FID across all experiments we perform.  This observation agrees with existing work \cite{rombach2021highresolution} and underscores the importance of choosing an appropriate $c$. We leave diving into the dynamic of latent channel dimension and downstream performance   as future work.

\begin{table}[t!]
\centering
\Large
\resizebox{0.9\linewidth}{!}{
\begin{tabular}{lcccccc}
\toprule
&$c$ &FID$\downarrow$   & sFID$\downarrow$  & IS$\uparrow$     & Precision$\uparrow$&Recall$\uparrow$ \\
\toprule
 & 4& 5.11&10.84&158.80&0.75&0.49\\
Fixed 16x   &8& {4.96}&\textbf{10.39}&\textbf{221.85}&\textbf{0.76}&\textbf{0.51}\\
& 16&\textbf{4.78}&11.81&187.47&0.72&0.49 \\
\midrule
 & 4& 5.12&11.12&152.39&0.72&0.48\\
 \vae & 8&\textbf{4.38}&\textbf{10.31}&181.03&\textbf{0.76}&0.48 \\
 &16& {4.56}&{10.55}&\textbf{191.09}&{0.75}&\textbf{0.49}\\
\bottomrule
\end{tabular}}
\caption{\textbf{Larger  channel $c$ is not always better for  generation.} Contrary to Table~\ref{tab:latentdim}, we find that increasing channel dimension does not always result in generation gains.}
\label{tab:latentdimgen}
\end{table}
\section{Discussion and Conclusion}

In this work, we propose an adaptive image tokenizer, \vae, which allocates different number of tokens to represent images based on content complexity derived from the text description of the image. Our experiments show that \vae improves both the quality and efficiency of image reconstruction and generation.
We identify several future directions to work on.
First,
we can apply  complexity-driven compression to developing discrete tokenizers and combine \vae with quantization techniques.
Besides, experimenting with more downstream tasks beyond class-conditional generation \citep{Shen_Yang_2021,shen2024scribeagent} and integrating \vae to  multi-modal models, such as Chameleon \cite{chameleonteam} and Transfusion \cite{transfusion}, can help strengthen this work.
Lastly, extending \vae to video tokenization could be a promising future direction due to the higher inherent redundancies in video clips, especially along the temporal dimension.

\section*{Acknowledgment}
We would like to thank Omer Levy, Daniel Li, Hu Xu for helpful discussion throughout the project.

\newpage
{
    \small
    \bibliographystyle{unsrtnat}
    \bibliography{main}
}

\clearpage

\setcounter{page}{1}

\maketitlesupplementary

\section{Prompt for LLM Scorer}
\label{sec:appen:prompt}
Our caption complexity pipeline works as follows:

\vspace{2mm}
Step 1: Use \texttt{Salesforce/instructblip-vicuna-7b} to generate caption, with the following prompts:

\begin{itemize}[leftmargin=.5in]
    \item What's in the image? $\rightarrow$ Caption
    \item Are there text or numbers in the image? $\rightarrow$ Yes/No.
    \item Are there faces in the image? $\rightarrow$ Yes/No.
\end{itemize}

\vspace{2mm}

Step 2: Use \texttt{meta-llama/Meta-Llama-3-70B-Instruct} to generate the complexity score with the prompt:
\vspace{2mm}

\fbox{
\parbox{0.9\textwidth}{
Given the description of a 512px image, determine its complexity based on the following factors:

1. Number of distinct objects

2. Color variance

3. Texture complexity

4. Foreground and background

5. Symmetry and repetition

6. Human perception factors, like the presence of human faces or text

You will be given the caption, whether there are text or numbers, and whether there are faces in the image. Assign a complexity score such that a higher number means the image is more complex. Note that text and facial details are intrinsically complex because they are crucial to human perception. Here are some examples for scoring:

- Score 1: A plane in a sky

- Score 2: A t-shirt with a emoji on it

- Score 3: A dog lying on the grass

- Score 4: A woman skiing in the snow

- Score 5: Two kids walking on the beach

- Score 6: A dinning table full of food

- Score 7: A close-up shot of a old man

- Score 8: Many people gathering in the stadium

- Score 9: Newspapers or graphs with text and numbers

Now determine the complexity for the caption:

[\textit{Insert caption here}]

[\textit{Insert one of  the following based on the Yes/No questions:}

\textit{- There are text visible in the image. There are also facial details.
}

\textit{- There are text visible in the image, but there is no human face.
}

\textit{- There is no obvious text in the image, but there are facial details.
}

\textit{- There is no text or human face in the image.
}]

Respond with ``Score: ? out of 9", where ``?" is a number between 1 and 9. Then provide explanations.

}
}

\section{Reconstruction Experiments}
\label{sec:appen:recon}
\subsection{Architecture}
\label{sec:appen:recon:arch}
We implement the nested VAE similar to the \texttt{AutoencoderKL} implementation of the \texttt{diffusers} library. The network configuration is:

\begin{itemize}[leftmargin=.5in]
    \item \texttt{sample\_size}: 512
    \item \texttt{in\_channels}: 3
    \item \texttt{out\_channels}: 3
    \item \texttt{down\_block\_types}: [\texttt{DownEncoderBlock2D}] $\times$ 6

    \item \texttt{up\_block\_types}: [\texttt{UpDecoderBlock2D}] $\times$ 6

    \item \texttt{block\_out\_channels}: [64, 128, 256, 256, 512, 512]
    \item \texttt{layers\_per\_block}: 2
    \item \texttt{act\_fn}: \texttt{silu}
    \item \texttt{latent\_channels}: 4/8/16
    \item \texttt{norm\_num\_groups}: 32
     \item \texttt{mid\_block\_attention\_head\_dim}: 1
    \item \texttt{num\_layers}: 8
\end{itemize}
The model sizes for different latent channels are shown below. As for the discriminator,  we use the pretrained StyleGAN~\cite{stylegan} architecture.
\begin{table}[h]
    \centering
    \small
    \begin{tabular}{lccc}
    \toprule
     Nested VAE  & $c=4$ & $c=8$ & $c=16$  \\
        \midrule
          \# Params (M)&187.45&187.50&187.61\\
        \bottomrule
    \end{tabular}
\end{table}

\subsection{Training}
\label{sec:appen:recon:train}
We use the following training configuration:

\begin{itemize}[leftmargin=.5in]
    \item GPU: 64 NVIDIA A100
    \item Per-GPU batch size: 8
    \item Global batch size: 512
    \item Training steps: 1,000,000
    \item Optimizer: AdamW
    \begin{itemize}
        \item \texttt{lr}: 0.0001
        \item \texttt{beta1}: 0.9
        \item \texttt{beta2}: 0.95
        \item \texttt{weight\_decay}: 0.1
        \item \texttt{epsilon}: 1e-8
        \item \texttt{gradient\_clip}: 5.0
    \end{itemize}
     \item {Scheduler}: {constant with 10,000 warmup steps }
    
    \item Loss:
    \begin{itemize}
    \item \texttt{recon\_loss\_weight}: 1.0
     \item \texttt{kl\_loss\_weight}: 1e-6
        \item \texttt{perceptual\_loss\_weight}: 1.0
        \item \texttt{moco\_loss\_weight}: 0.2
       
        \item \texttt{gan\_loss\_weight}: 0.5
        \item \texttt{gan\_loss\_starting\_step}: 50,000
         
    \end{itemize}
    
\end{itemize}

The discriminator is trained with the standard GAN loss.

\subsection{Baselines}
\label{sec:appen:recon:baseline}

\begin{wrapfigure}{r}{0.43\textwidth}
\centering
\vspace{-1.6cm}
    \includegraphics[width=0.43\textwidth]{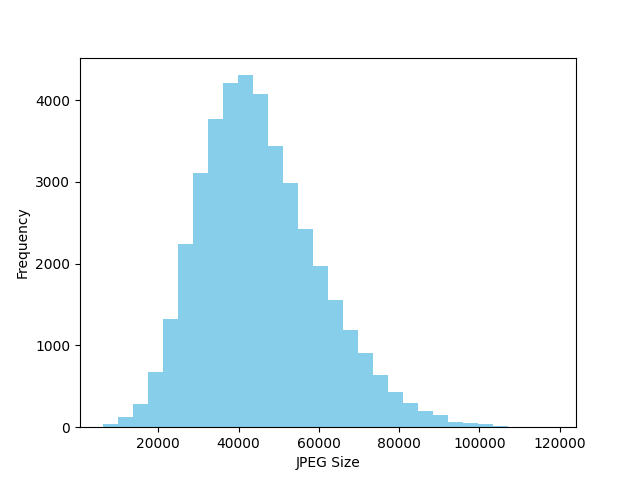}
    \vspace{-0.8cm}
    \caption{On COCO 2014 test set, the minimum JPEG size is 6128; maximum is 118428; mean is 45474.29; standard deviation is 15037.07.}
    \label{fig:jpeg_dist}
        \vspace{-0.8cm}
\end{wrapfigure}
We train fixed compression baselines using the same  data, training configuration, and VAE backbone. For smaller compression ratios, e.g., fixed 8x, the last two downsampling blocks and first two upsampling blocks are not used.

For the adaptive JPEG baseline, 
we use \texttt{torchvision.io.encode\_jpeg} to transform the images into JPEG file and compute the number of bytes as the complexity metric. 
Smaller files correspond to larger complexity. To provide a better understanding of this metric, we show in Figure~\ref{fig:jpeg_dist} the distribution of JPEG sizes on the COCO 2014 test set, with relevant statistics included in the caption. Then, based on the JPEG sizes of all images in the Shutterstock training dataset, we set the thresholds $(a,b)$ to $(38761, 65837)$ to categorize the file sizes into three compression ratios. This set of thresholds ensure that the JPEG baseline has the same training compression ratio distribution as \vae. 

For LDM VAEs, we follow the instructions in their original repository to use the model checkpoints. Note that LDM VAEs are trained on OpenImages dataset \cite{openimages}, which is different from our training data, so it is hard to fairly compare the reconstruction performance. Nonetheless, we present their rFIDs on the evaluation datasets in Table~\ref{tab:ldmcompare}.

\begin{table}[h!]
\centering
\small
\begin{tabular}{lcccc}
\toprule
& {COCO}  &{ImageNet}&  {CelebA} & {ChartQA}\\
\toprule
 \vae&0.65&0.46&1.97&5.27\\
 LDM 8x&0.51&0.33&2.83&8.32 \\

 LDM 16x&0.53&0.37&  3.07 &8.49 \\
 LDM 32x&0.90&0.62&5.54&10.35\\

\bottomrule
\end{tabular}
\caption{rFIDs for \vae and LDM VAEs.}
\label{tab:ldmcompare}
\end{table}

\subsection{More Reconstruction Visualization}
\label{sec:appen:recon:visual}
\begin{figure*}[t!]
    \centering
    \includegraphics[width=0.9\linewidth]{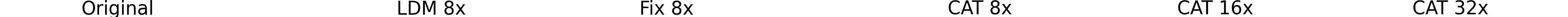}
    \includegraphics[width=0.9\linewidth]{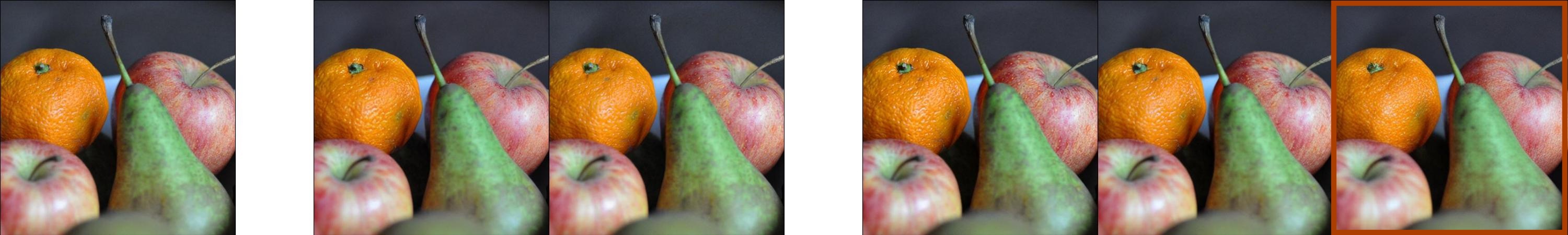}
    \includegraphics[width=0.9\linewidth]{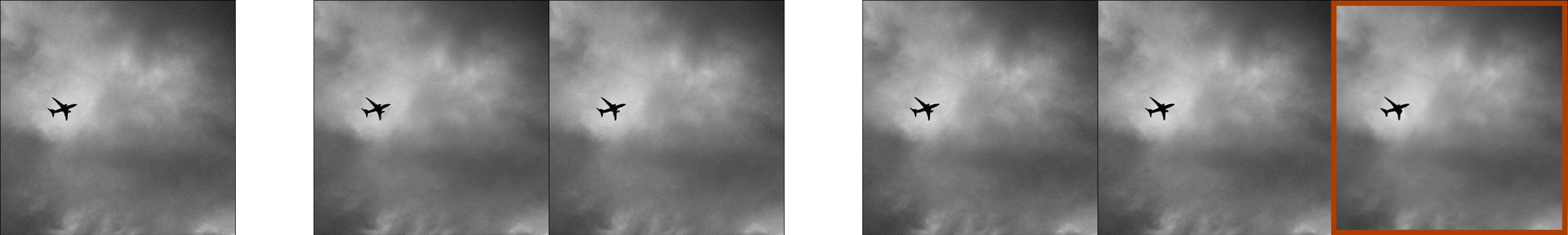}
    \includegraphics[width=0.9\linewidth]{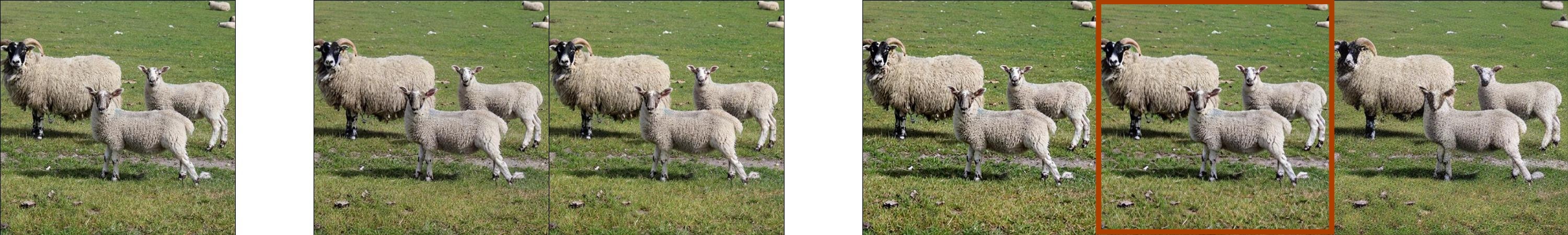}

    \includegraphics[width=0.9\linewidth]{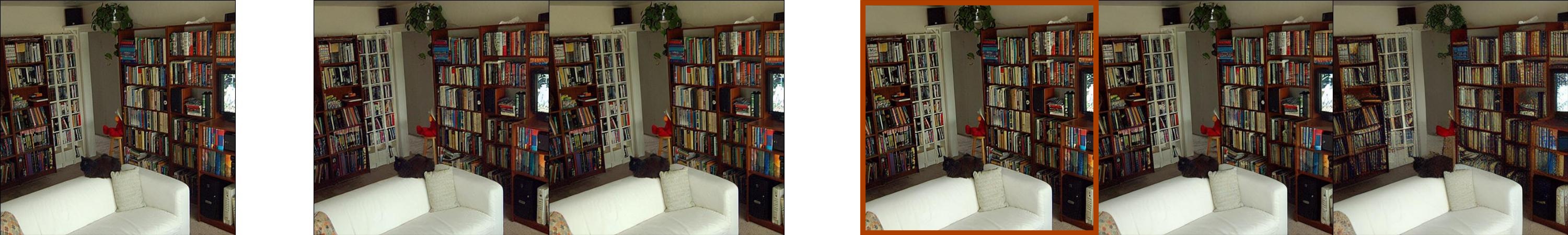}

    \includegraphics[width=0.9\linewidth]{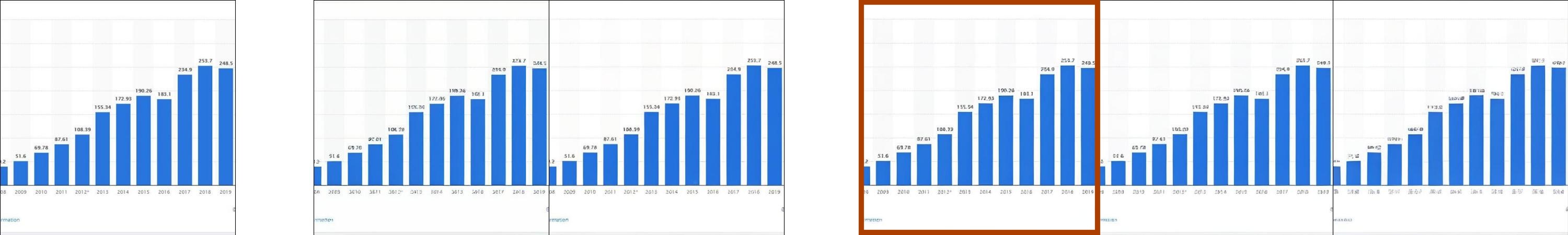}
    \includegraphics[width=0.9\linewidth]{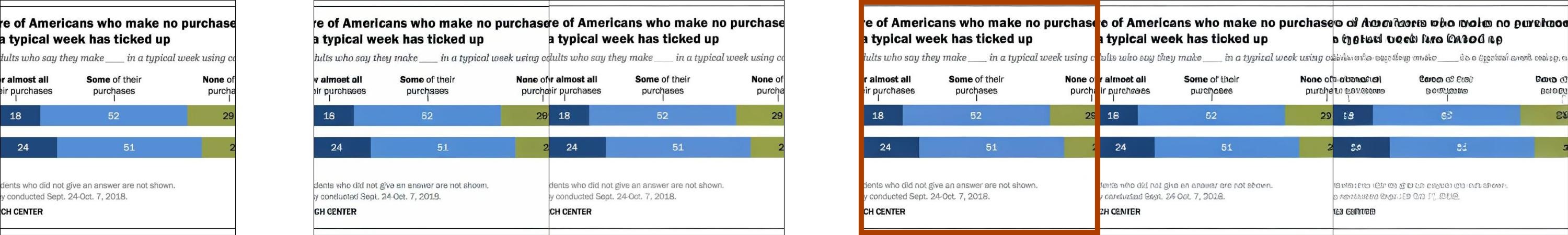}
    \includegraphics[width=0.9\linewidth]{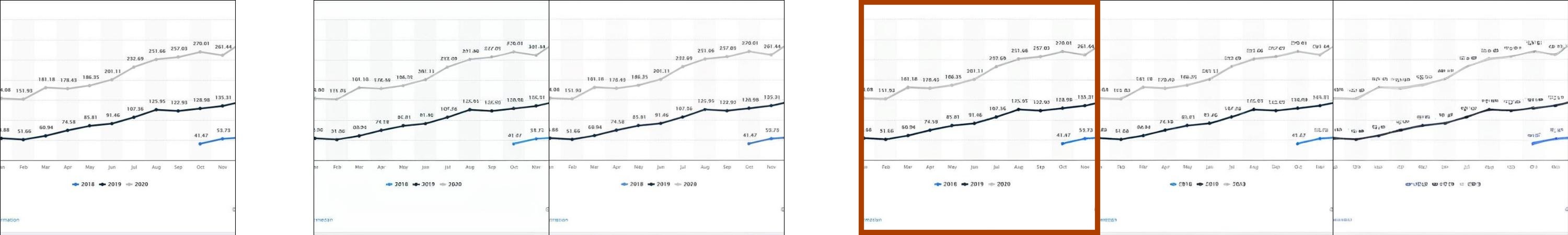}
    \includegraphics[width=0.9\linewidth]{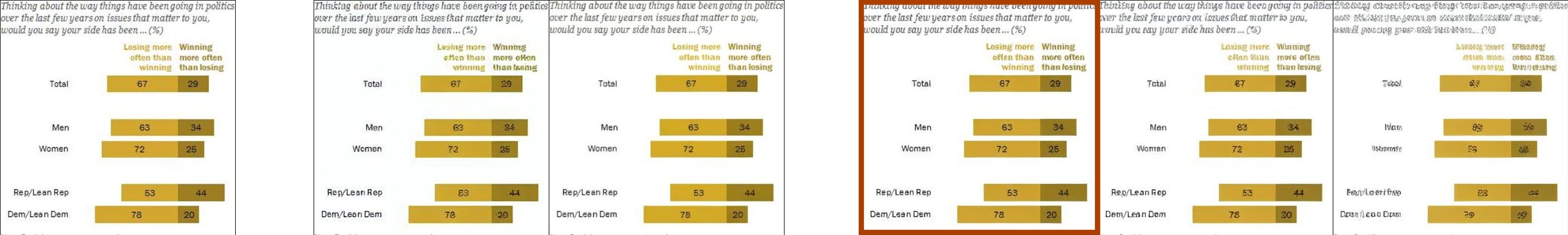}
    \vspace{-2mm}
    \caption{\textbf{More \vae reconstruction examples. We highlight the compression ratio selected by our proposed caption complexity in red.} Images shown in the figure are taken from COCO 2014 \citep{coco} and ChartQA \citep{chartqa}. }
    \label{fig:appen:reconvisual}
\end{figure*}

See Figure~\ref{fig:appen:reconvisual} in the end.

\section{Generation Experiments}
\label{sec:appen:gen}
\subsection{Architecture}
\label{sec:appen:gen:arch}
We use DiT-XL architecture with a patchify downsampler and  patch size of 2. The model size depends on the latent channel, but is generally around 431M parameters. The model Tflops is 22.0.

\subsection{Training \& Inference}
\label{sec:appen:gen:train}
The training configuration for DiT is as follows:
\begin{itemize}[leftmargin=.5in]
    \item GPU: 16 NVIDIA H100
    \item Per-GPU token batch size: 4096 $\times$ 4  (equivalent to 64 images for 16x compression ratio)
    \item Global token batch size: 4096 $\times$ 64
    \item Training steps: 400,000
    \item Optimizer: AdamW
    \begin{itemize}
    \item \texttt{lr}: 0.0001
        \item \texttt{beta1}: 0.9
        \item \texttt{beta2}: 0.95
        \item \texttt{weight\_decay}: 0.1
         \item \texttt{epsilon}: 1e-8
        \item \texttt{gradient\_clip}: 1.0
    \end{itemize}
    \item {Scheduler}: {Cosine}
     \begin{itemize}
     \item \texttt{warmup}: 4000
        \item \texttt{cosine\_theta}: 1.0
        \item \texttt{cycle\_length}: 1.0        
        \item \texttt{lr\_min\_ratio}: 0.05        
    \end{itemize}
\end{itemize}

\vspace{2mm}
\noindent
DDPM scheduler (\texttt{diffusers} implementation):
\begin{itemize}[leftmargin=.5in]
\item \texttt{num\_train\_timesteps}: 1000
            \item \texttt{beta\_start}: 0.0001 
            \item \texttt{beta\_end}: 0.02
        \item \texttt{beta\_schedule}: \texttt{squaredcos\_cap\_v2}
        \item \texttt{prediction\_type}: \texttt{epsilon}
        \item \texttt{timestep\_spacing}: \texttt{leading}
       
        \item \texttt{num\_inference\_steps}: 250        
    \end{itemize}
For 10 \% of the time, we remove the image class label from the input and train unconditional image generation. All FID-50K and images generated in this paper are using cfg=1.5.

\subsection{Baselines}
\label{sec:appen:gen:baseline}
To ensure we train the baseline with the same compute FLOPs, we fix the token batch size and number of training steps for all settings. 
For pretrained LDM VAE, we use the scaling factor specified in the model configuration to ensure the input scale and noise scale are similar. For \vae, we use a scaling factor of 1.

\subsection{More Visualization}
\label{sec:appen:gen:visual}
\begin{figure*}
    \centering
    \includegraphics[width=0.9\linewidth]{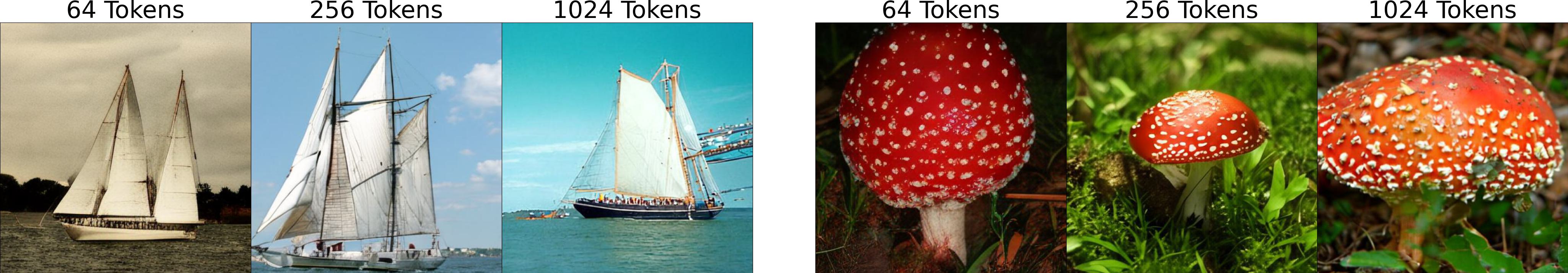}
    \includegraphics[width=0.9\linewidth]{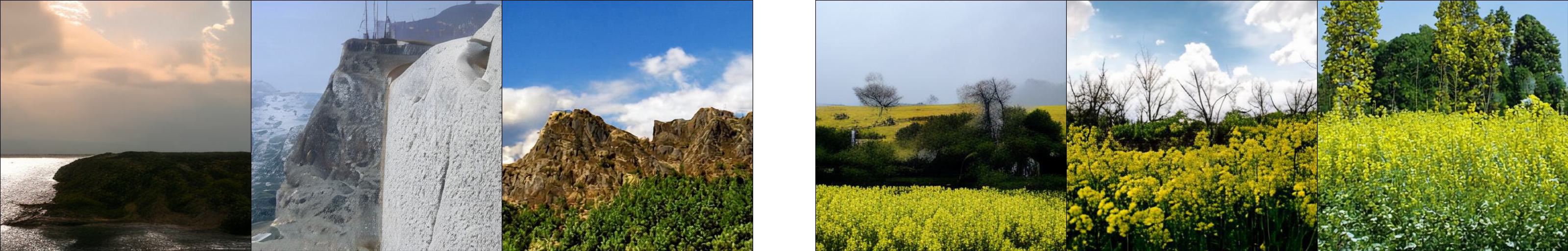}
    \includegraphics[width=0.9\linewidth]{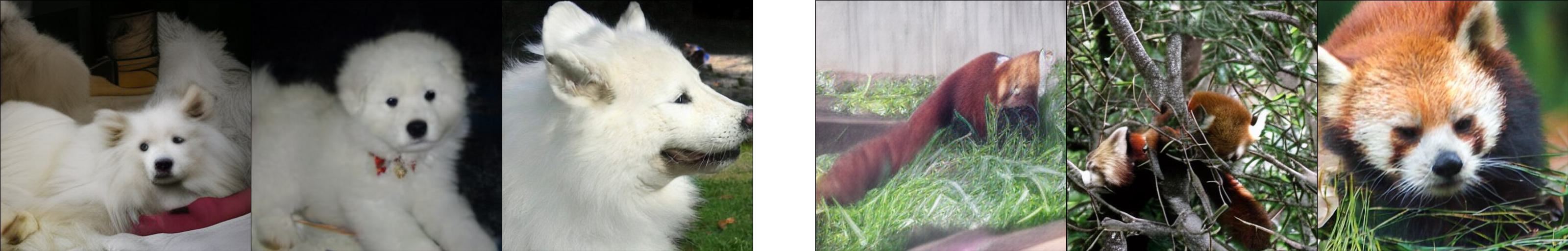}
    \includegraphics[width=0.9\linewidth]{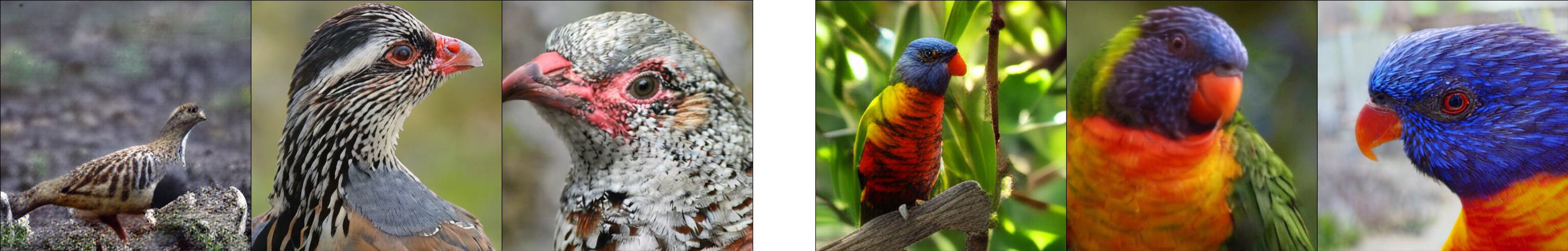}
    \includegraphics[width=0.9\linewidth]{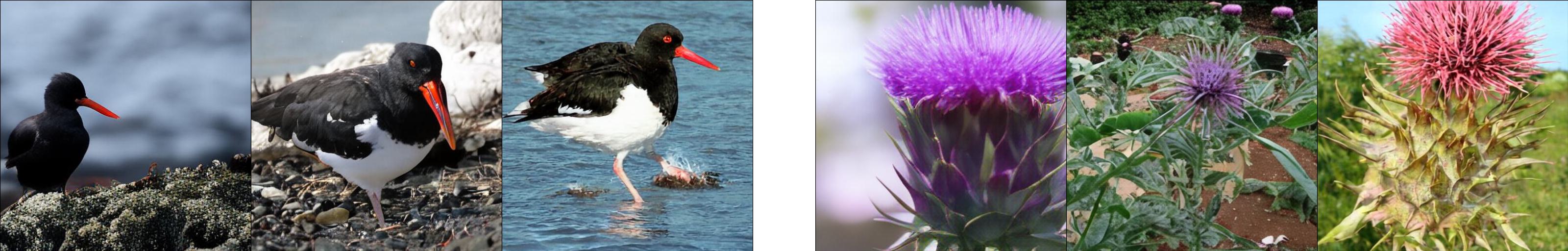}
    \includegraphics[width=0.9\linewidth]{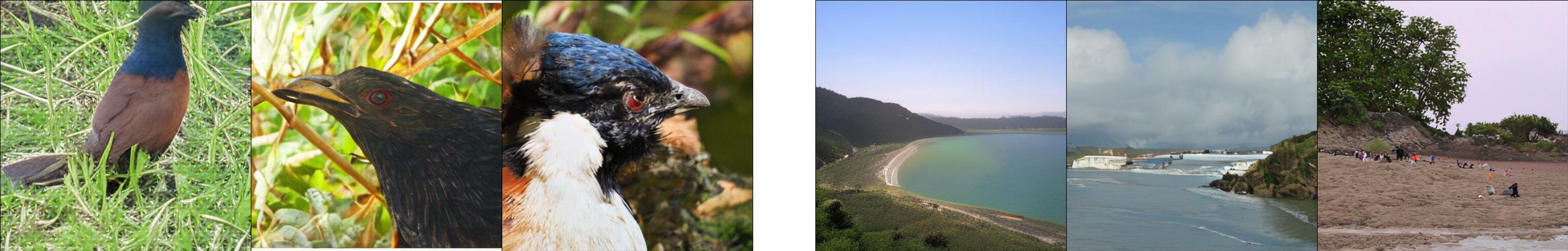}
    \includegraphics[width=0.9\linewidth]{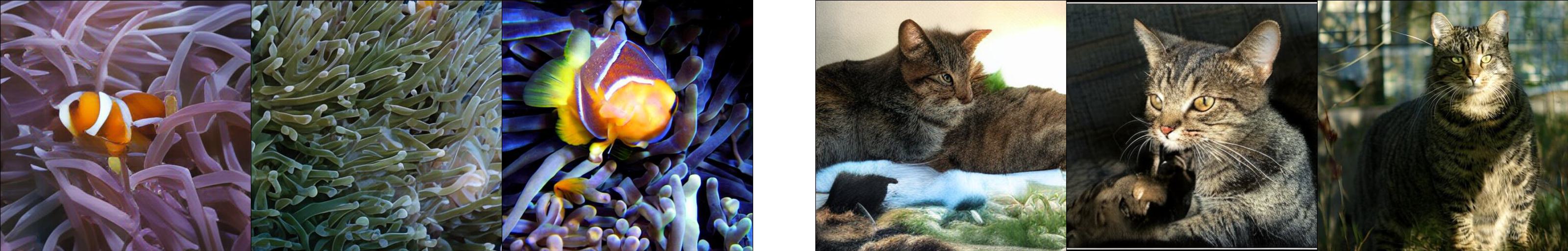}
    \caption{\textbf{More DiT-\vae generation examples. Increasing token count (left$\rightarrow$right) generally  leads to better image quality and higher complexity.}}
    % \cz{the 1024 elephant and 1024 ship don't look good. 1024 building is okay but not very good.}}
    \label{fig:appen:genvisual}
\end{figure*}

See Figure~\ref{fig:appen:genvisual} in the end.

% \oldreconstructiontable

\end{document}